\documentclass[conference]{IEEEtran}

\usepackage[numbers]{natbib}
\usepackage{multicol}
\usepackage[bookmarks=true]{hyperref}

\ifCLASSINFOpdf
  \usepackage[pdftex]{graphicx}
  \graphicspath{{./pdf/}{./jpeg/}}
  \DeclareGraphicsExtensions{.pdf,.jpeg,.PNG}
\else
  \usepackage[dvips]{graphicx}
  \graphicspath{{./eps/}}
  \DeclareGraphicsExtensions{.eps}
\fi

\usepackage{xcolor}
\usepackage[switch]{lineno}
\usepackage{amsmath,amssymb,amsthm,mathtools}
\usepackage{algorithm}
\usepackage{algorithmic}

\usepackage{bm}
\usepackage{multirow}
\usepackage{makecell}

\usepackage{graphics}
\usepackage[font=small]{caption}
\usepackage{color, soul}
\usepackage{subcaption}
\usepackage{amsfonts}
\usepackage{amsmath}
\usepackage{amssymb}
\usepackage{mathtools}
\usepackage{diagbox}
\usepackage{siunitx}
\usepackage{subcaption}
\usepackage{tikz}
\usepackage[utf8]{inputenc}
\begin{document}

\title{Global-Local Interface for On-Demand Teleoperation}

\author{Jianshu Zhou$^{*}$,
        Boyuan Liang$^{*}$, 
        Junda Huang,
        Ian Zhang,
        Masayoshi Tomizuka
        \\
        \\University of California, Berkeley}

\twocolumn[{%
    \renewcommand\twocolumn[1][]{#1}%
    \maketitle
    \begin{center}
        \centering
        \includegraphics[width=0.95\textwidth]{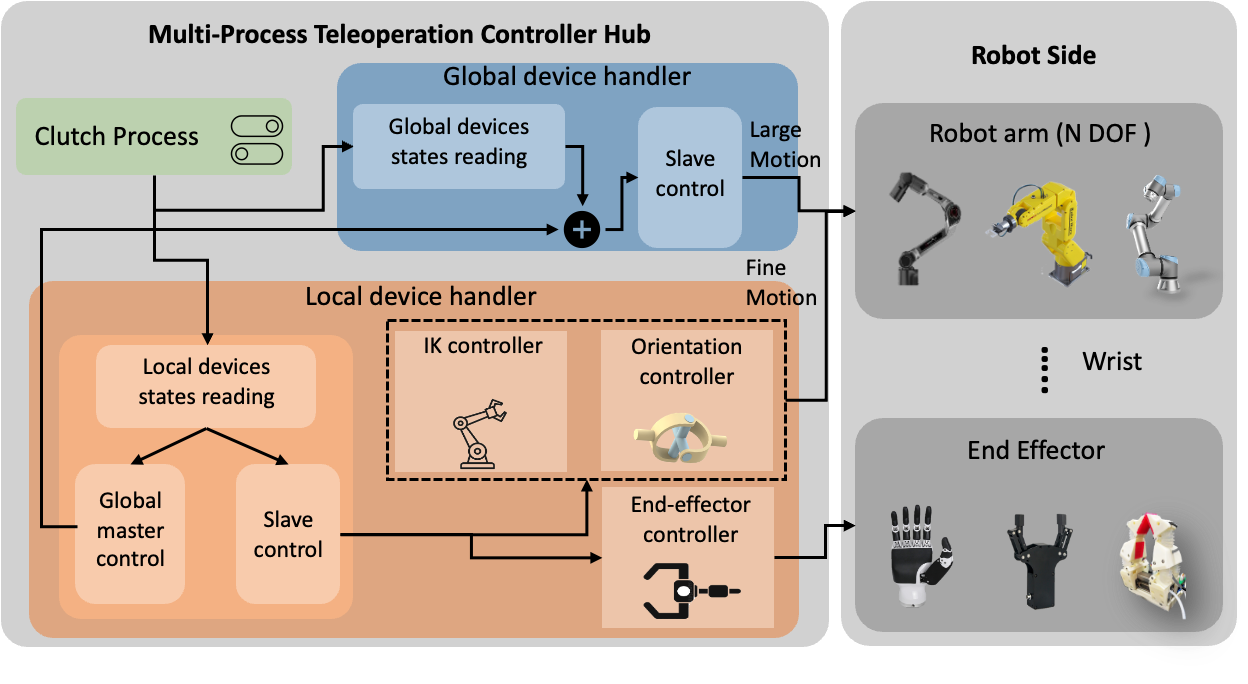} %
    \captionof{figure}{Overview of the Global-Local Teleoperation Interface. The master system is decomposed into the global and local component. The global component controls rapid large-range motions and reflect the general collision configuration of the slave system. The local component controls the accurate or dexterous local motions. Their detailed realizations, and whether they are allowed to be activated simultaneously, can be adapted based on task requirements.}
    \label{fig:overview}
    \end{center}%
    }]

\thispagestyle{empty}
\pagestyle{empty}

\begin{abstract}
Teleoperation is a critical method for human-robot interface, holds significant potential for enabling robotic applications in industrial and unstructured environments. Existing teleoperation methods have distinct strengths and limitations in flexibility, range of workspace and precision. To fuse these advantages, we introduce the Global-Local (G-L) Teleoperation Interface. This interface decouples robotic teleoperation into global behavior, which ensures the robot motion range and intuitiveness, and local behavior, which enhances human operator's dexterity and capability for performing fine tasks. The G-L interface enables efficient teleoperation not only for conventional tasks like pick-and-place, but also for challenging fine manipulation and large-scale movements. Based on the G-L interface, we constructed a single-arm and a dual-arm teleoperation system with different remote control devices, then demonstrated tasks requiring large motion range, precise manipulation or dexterous end-effector control. Extensive experiments validated the user-friendliness, accuracy, and generalizability of the proposed interface.
\end{abstract}


\section{Introduction}
\label{sec:intro}

\begin{table*}[ht]
\centering
\resizebox{\textwidth}{!}{%
\begin{tabular}{|>{\centering\arraybackslash}m{3cm}|
                >{\centering\arraybackslash}m{3cm}|
                >{\centering\arraybackslash}m{3cm}|
                >{\centering\arraybackslash}m{3cm}|
                >{\centering\arraybackslash}m{3cm}|
                >{\centering\arraybackslash}m{3cm}|}
\hline
\textbf{Feature} & \textbf{G-L Interface (Bi)} & \textbf{G-L Interface (Si)} & \textbf{Gello} \cite{wu2023gello} & \textbf{ACE}\cite{yang2024ace} & \textbf{OpenTelevision} \cite{cheng2024open} \\ \hline \hline
\textbf{End-effector Type} & Two-fingered Gripper & Dexterous Hand & Two-Fingered Gripper & Gripper and dexterous hand & Dexterous hand \\ \hline
\textbf{Arm Mode} & Bimanual & Single & Bimanual & Bimanual & Bimanual \\ \hline
\textbf{Master Device} & Arm replica + Touch X & Arm replica + IMU + exoskeleton & Arm Replica & Arm replica + RGBD camera & VR (VisionPro) \\ \hline
\textbf{Teleoperation of Arm} & Joint mapping + Inverse Kinematics & Joint mapping + Euler angle& Joint mapping & Joint mapping & Inverse kinematics \\ \hline
\textbf{Teleoperation of End-effector} & Boolean button & Joint mapping & Joint mapping & Hand tracking & Hand tracking \\ \hline \hline
\textbf{Motion Range} & Full joint space & Full joint space & Partial joint space & Front cartesian space & Front cartesian space \\ \hline
\textbf{In-device collision} & None & None & Restricted & None & None \\ \hline
\textbf{Precision} & High & Mid & Mid & Mid & Mid \\ \hline
\end{tabular}%
}
\caption{Comparison of Teleoperation Interfaces}
\label{tab:teleoperation}
\end{table*}

With the advancement of embodied intelligence, there is an increasing expectation for smarter robots capable of operating across a broader range of scenarios, including industrial settings and unstructured environments \cite{billard2019trends, rus2015design}. Teleoperation has gained growing recognition as a critical human-robot interface \cite{vertut2013teleoperation, darvish2023teleoperation, slade2024human}. Teleoperation has demonstrated exceptional value across various domains and scenarios, such as robotic surgery, where precise movements and real-time decision-making are indispensable \cite{pugin2011history, guo2019scaled, haidegger2011surgery}; underwater exploration, which grants access to hazardous and challenging environments \cite{brantner2021controlling}; space exploration, extending human reach beyond Earth's boundaries for scientific discoveries and exploration \cite{sheridan1993space, yoon2004model}; and disaster response, providing essential assistance in dangerous situations, such as search and rescue operations \cite{katyal2014approaches, cardenas2019design,ramos2019dynamic}. In addition, teleoperation offers efficient and cost-effective means for collecting high-quality demonstration data \cite{wu2023gello, darvish2023teleoperation}. This is crucial for advancing robotic learning frameworks, such as imitation learning \cite{fu2024mobile, chi2023diffusion,Huang2025DIHtele} and vision-language-action (VLA) models \cite{zitkovich2023rt, chi2024eva}, where data availability has become a significant bottleneck \cite{argall2009survey, zhang2018deep, vuong2023open}.

Although various teleoperation approaches have been proposed, each exhibits distinct strengths and limitations. Some typical methods utilize a remote floating device, representing the master's end-effector, and an external tracking system, often vision-based, to determine its pose \cite{selvaggio2021shared, dass2024telemoma, iyer2024open,cheng2024open,ding2024bunny,park2024dexhub,qin2023anyteleop, lin2024learning, yang2024ace, zhang2018deep}. Under this framework, the master system almost always has six degrees of freedom (DoF), representing the operator's hand position and orientation. While this approach facilitates greater operator dexterity due to the absence of mechanical constraints, challenges arise in both aligning the workspaces of the master and slave systems and mitigating the loss in the super-human's working space intrinsic to the robot arm. Even when the master and slave possess the same number of DoF, discrepancies in their workspace geometries can lead to situations where the operator specifies unattainable poses for the slave, or difficulty for the operator to drive the slave system to a reachable point. This issue is further compounded by differences in the number of DoF. When the slave possesses fewer than six DoF, the likelihood of specifying unattainable poses increases. Conversely, for slaves with more than six DoF, the framework cannot adequately control the slave's nullspace, hindering the operator's capability to leverage redundant DoF for more complicated tasks such as collision avoidance.

On the other hand, employing a scaled replica of the slave system as the master device is also a popular design. This framework provides the operator with an intuitive understanding of the slave robot's workspace, including reachability and potential collision configurations, due to the preserved kinematic structure \cite{kumar2019review, tran2021hand, wu2023gello, fu2024mobile, aldaco2024aloha, fang2024airexo}. However, the inherent mechanical constraints, teleoperation redundancy, and detent torque in encoders of the master system can impose unwanted movements or apply varying external forces which can significantly hinder the operator's dexterity and capability to perform fine-grained tasks.

Recognizing the limitations of existing teleoperation frameworks, we propose the novel Global-Local (G-L) Teleoperation Interface. The G-L interface addresses teleoperation challenges by decoupling the master system into global and local components. The global component facilitates efficient navigation and reflects the collision configurations of the slave system within its workspace. This component enables rapid, large-scale adjustments of the slave’s pose, ensuring comprehensive workspace coverage. The local component focuses on fine-grained control of the slave’s end-effector, delivering the dexterity and precision required for intricate tasks. By integrating these complementary components, the G-L Teleoperation Interface fully leverages the slave system's capabilities, overcoming human limitations while preserving the operator’s precision and dexterity, and ultimately enables the execution of complex tasks requiring both broad reach and fine manipulation. Fig \ref{fig:overview} shows a schematic diagram of how G-L interface operates. The experiments demonstrate the enhanced dexterity, precision, and motion range of two specialized task-specific teleoperation systems under the G-L interface.

\section{Global-Local Teleoperation Interface}
\label{sec:method}

This section details the decoupling of master systems into global and local components within the G-L Teleoperation Interface. To be considered a valid implementation of the G-L Interface, a decoupled master system must adhere to the following criteria. First, the global component must provide an intuitive and comprehensive representation of the slave system's collision configuration. It should enable rapid, large-amplitude adjustments of the slave's pose, prioritizing efficient workspace navigation over fine manipulation or dexterity. Second, the local component must facilitate precise, fine-grained control of the slave's end-effector, enabling the operator to execute tasks requiring dexterity and accuracy. It should prioritize the seamless transfer of the operator's manipulation skills to the slave system, with less emphasis on reflecting the slave's overall configuration or controlling all degrees of freedom (DOFs). Finally, the combined functionality of the global and local components must ensure complete control over all DOFs of the slave system.

This work mainly explores two decoupling methods for realizing the G-L interface which we have termed as temporal and spatial decoupling. In temporal decoupling, both global and local components possess full control over all slave DOFs. However, they are activated sequentially. The operator first utilizes the global component for gross positioning of the slave system, followed by the local component for fine adjustments to achieve the desired task. Spatial decoupling divides the slave DOFs into two groups. The global component controls the first group, primarily influencing end-effector position and overall slave collision configuration. The local component governs the second group, primarily determining end-effector orientation with minor impact on the overall configuration. The global and local components can be used simultaneously since they control different DOFs. It is important to note that these methods are not exhaustive. The G-L interface accommodates various decoupling strategies tailored to specific slave systems and task demands.

The subsequent subsections elaborate on the design and implementation of temporal and spatial decoupling. Section \ref{sec:setup} will showcase their physical realizations.

\subsection{Temporal Decoupling}
\label{subsec:temporal-decoup}

\begin{figure}[!t]
    \centering
    \includegraphics[width=\linewidth]{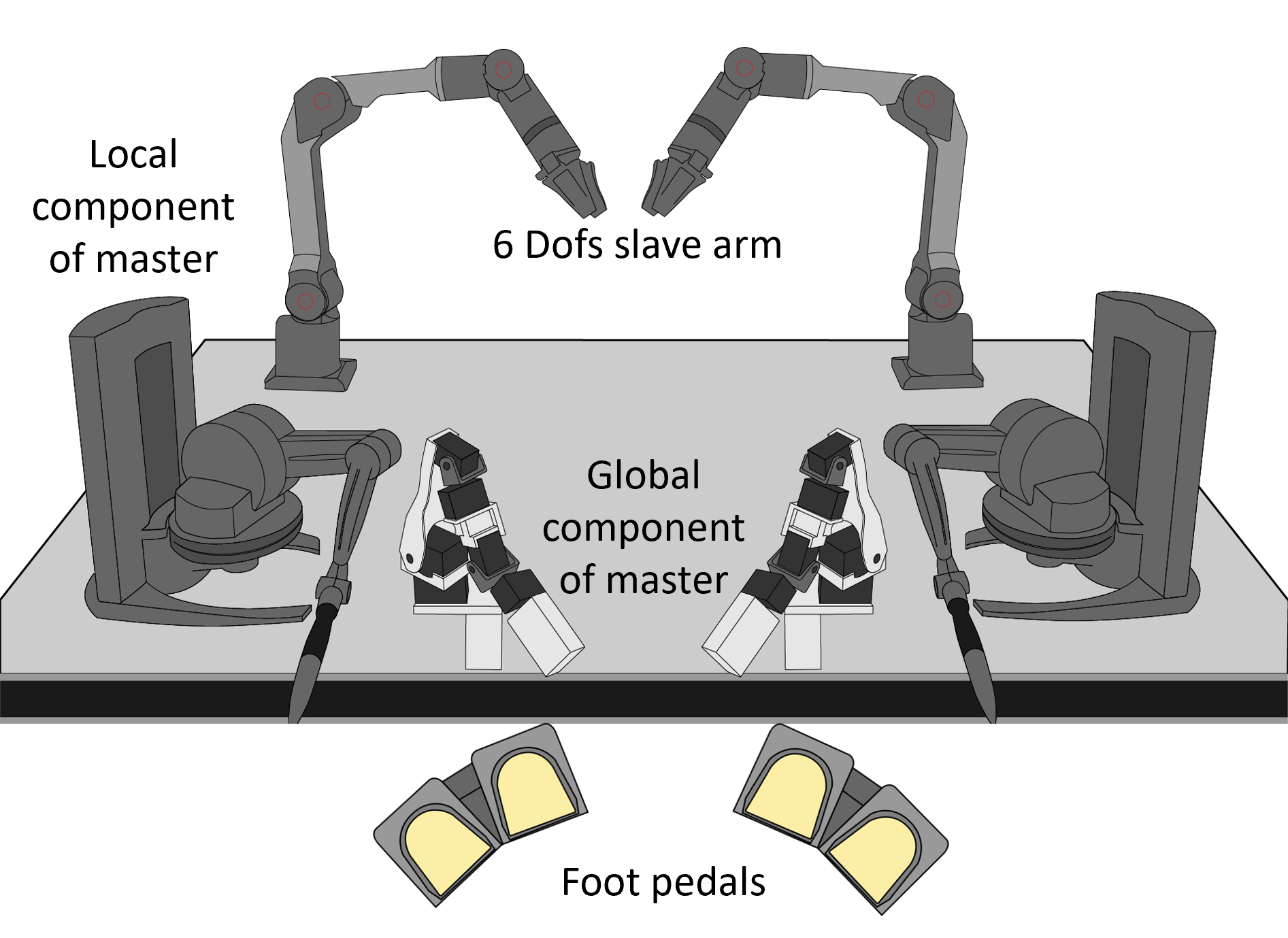}
    \caption{A schematic diagram of temporal decoupling. The global component is a scaled replica of the slave robot and the local component is a haptic device. The two step pedals allows the operator to rapidly switch between the local and global component. If the local component is activated, the global component will temporary turn into a slave system that follows the actual slave system to avoid joint position mismatch when switching back to the global component.}
    \label{fig:temporal-method}
\end{figure}

In the temporal decoupling scheme, the global component is realized as a scaled replica of the slave system, a common approach in teleoperation \cite{fu2024mobile, aldaco2024aloha}. The local component utilizes a haptic device for precise tracking of the operator's fingertip position and orientation. Activating the global component places the slave system in joint control mode, replicating the scaled replica's joint positions within the limits of the motor capacities. Conversely, activating the local component switches the slave system to end-effector Cartesian control, where it tracks the motion of the haptic device. Fig. \ref{fig:temporal-method} shows a schematic diagram of temporal decoupling.

To facilitate precise control of the end-effector, the slave system is designed to track only a portion of the local component motion. Every time the local component is activated, its current position $p_0$ and orientation $q_0$ will be re-initialized as the origin. Given a linear scaling factor $\alpha_l\in(0,1]$, the slave system will move its end-effector position towards the scaled linear displacement of the haptic device, $\alpha_l(p-p_0)$. For rotational motions, assume the orientation of the local component has moved from $q_0$ to $q$, we convert the rotational displacement $q_0^{-1}q$ into axis-angle representation $(\vec{v},\theta)$, where $\vec{v}$ is the rotational axis and $\theta$ is the rotational angle. Assume the slave system's end-effector currently has orientation $q_e$, given a rotational scaling factor $\alpha_r\in(0,1]$, the slave system will target to change its end-effector orientation via rotating around axis $\vec{v}$ by angle $\alpha_r\theta$, starting from $q_e$.

This design presents a challenge during the transition from the local to global component. A mismatch can arise between the pose of the scaled replica (global component) and the actual slave system configuration, potentially causing an abrupt motion in the slave upon returning to global control. To mitigate this, the global component assumes a "slave" role during local component activation, mirroring the joint motions of the true slave system. This ensures pose synchronization between the global component and the slave, facilitating a smooth transition when switching back to global controls.

It is worth noting that there is an overlap between the DOFs that the global and local components are controlling, so they cannot be activated at the same time. Using devices like foot pedals allows the operator to conveniently switch between the global and local components.

\subsection{Spatial Decoupling}
\label{subsec:spatial-decoup}

\begin{figure}[!t]
    \centering
    \includegraphics[width=\linewidth]{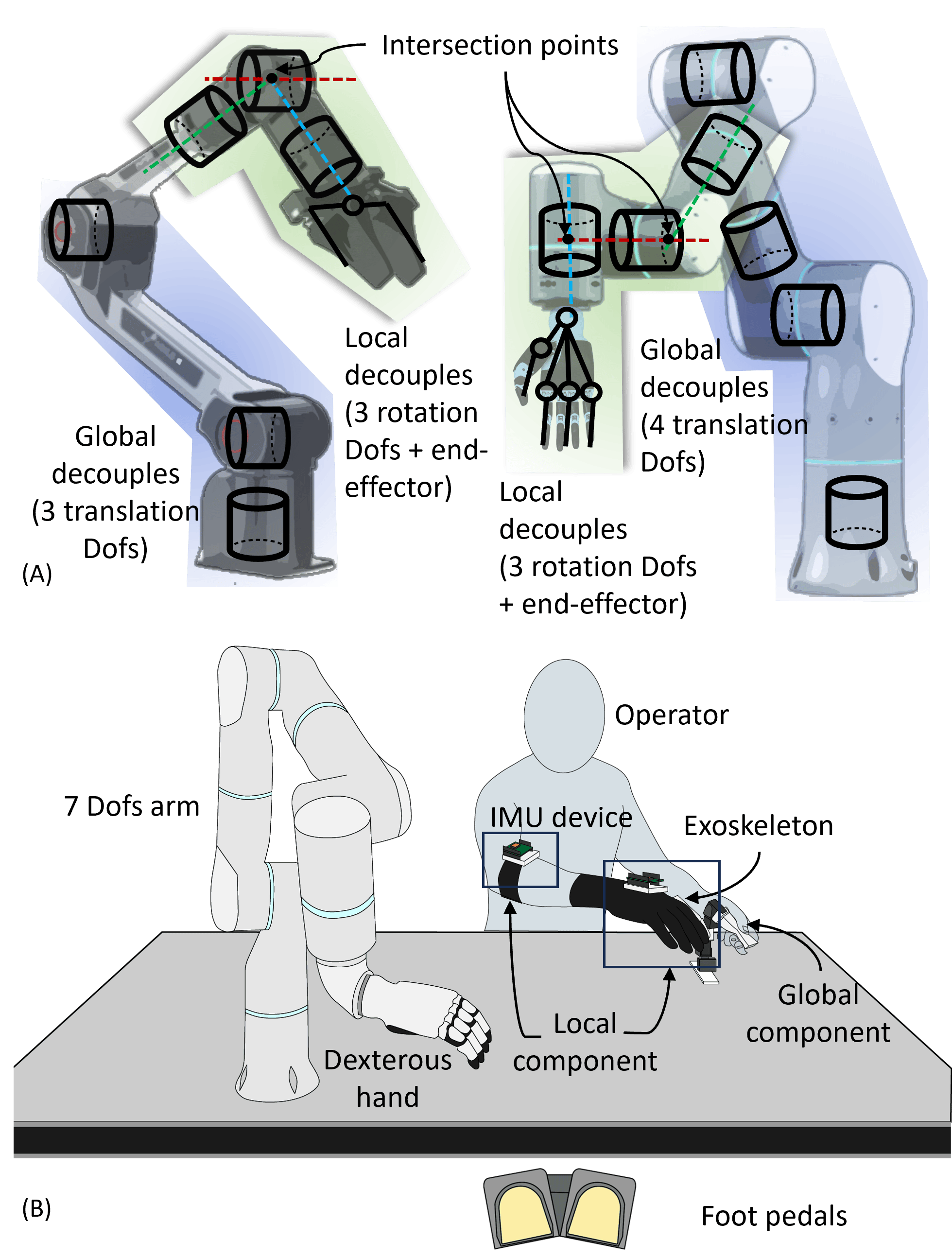}
    \caption{A schematic diagram of spatial decoupling. The global component is a scaled replcia of the first $N-3$ DOFs, while the local component consists of two IMU sensors to measure human operator's wrist rotation and map it back to the last three joints's target positions. Since it eliminates all mechanical constraints around the operator's hand, it allows mounting complicated end-effector controllers, such as an exoskeleton structure for controlling a dexterous hand.}
    \label{fig:spatial-method}
\end{figure}

Spatial decoupling leverages a common characteristic of serially-linked industrial robots in that the final three joints primarily influence end-effector orientation with less impact on the overall robot configuration, while the remaining joints predominantly determine end-effector position and potential collisions.  This principle is applied to an $N$-DOF industrial robot (e.g., KUKA or FANUC) by dividing control into two components. The global component replicates the first $N-3$ joints, providing gross positioning, while a local component, consisting of a dual-IMU system, mounted on the forearm and the hand of the human operator. measures the relative rotation $R$ between the human wrist and hand to control the final three joints. This approach often simplifies inverse kinematics to finding X-Y-Z or X-Y-X Euler angles based on $R$. Fig. \ref{fig:spatial-method} shows a schematic diagram of spatial decoupling.

This decoupling architecture allows simultaneous activation of both components. The operator can manipulate the global component with one hand and the local component with the other. While seemingly adding complexity, this approach offers two key advantages. First, the dual-IMU design in the local component eliminates mechanical constraints around the operator's hand, allowing integration of additional devices to control more sophisticated end-effectors. Second, replicating the first $N-3$ joints enables full control of the slave robot's redundant DOFs, maximizing workspace utilization. Existing teleoperation systems often prioritize one of these advantages. Systems focusing on precise hand pose tracking \cite{cheng2024open,yang2024ace,qin2023anyteleop} may have limited control over the full robot workspace. Conversely, systems emphasizing workspace utilization \cite{fu2024mobile, aldaco2024aloha, wu2023gello} may impose constraints on the operator's hand, potentially hindering dexterity and the use of additional tools. Spatial decoupling combines both advantages, as demonstrated in the experiment section where a glove-shaped device is mounted on the local component to teleoperate a multi-fingered hand.

\begin{figure}[!t]
    \centering
    \includegraphics[width=\linewidth]{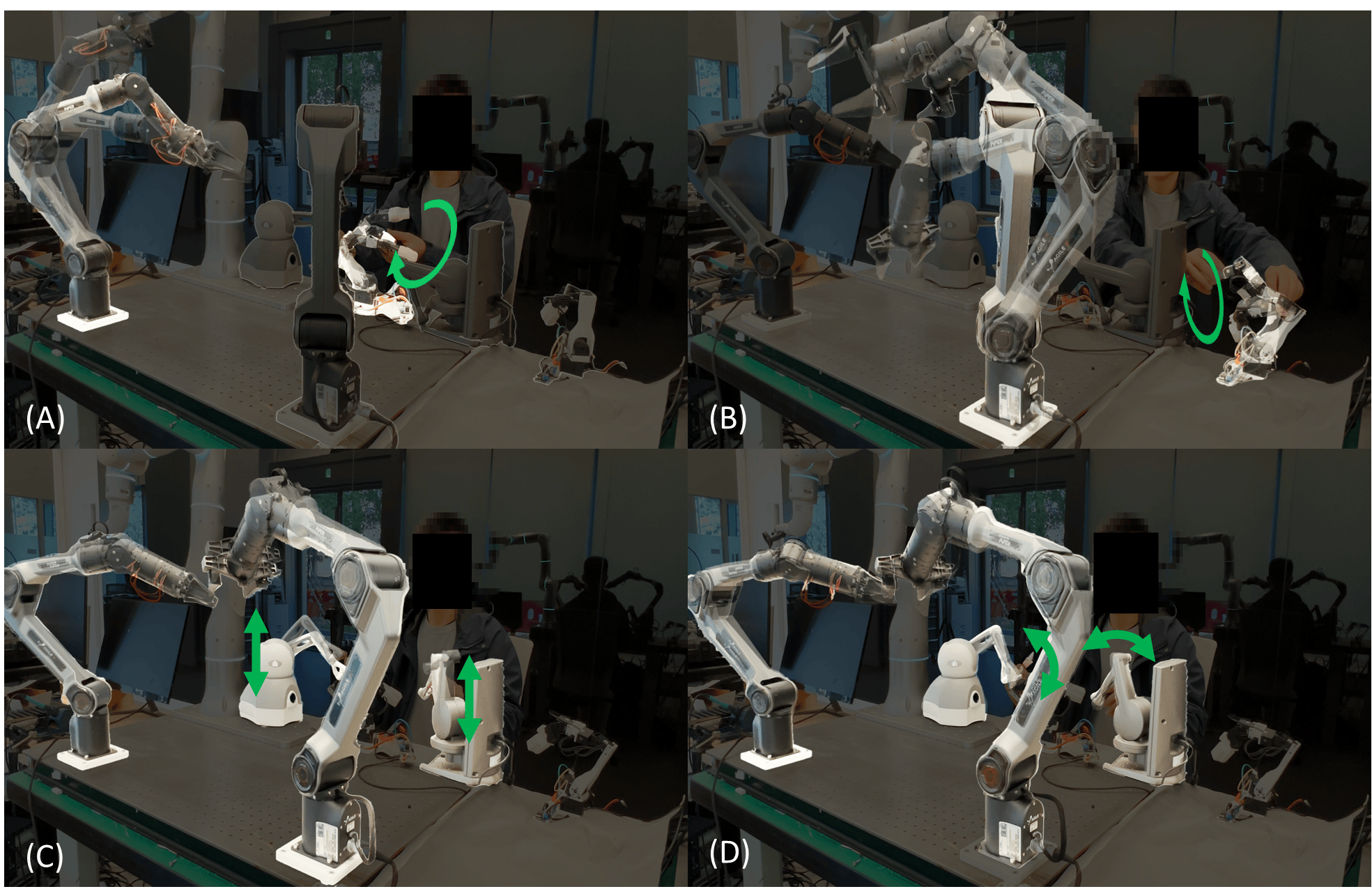}
    \caption{Physical realization of temporal decoupling. (A), (B) Teleoperation with global component. (C), (D) Teleoperation with local component.}
    \label{fig:implement-temporal}
\end{figure}

\begin{figure}[!t] 
    \centering
    \includegraphics[width=\linewidth]{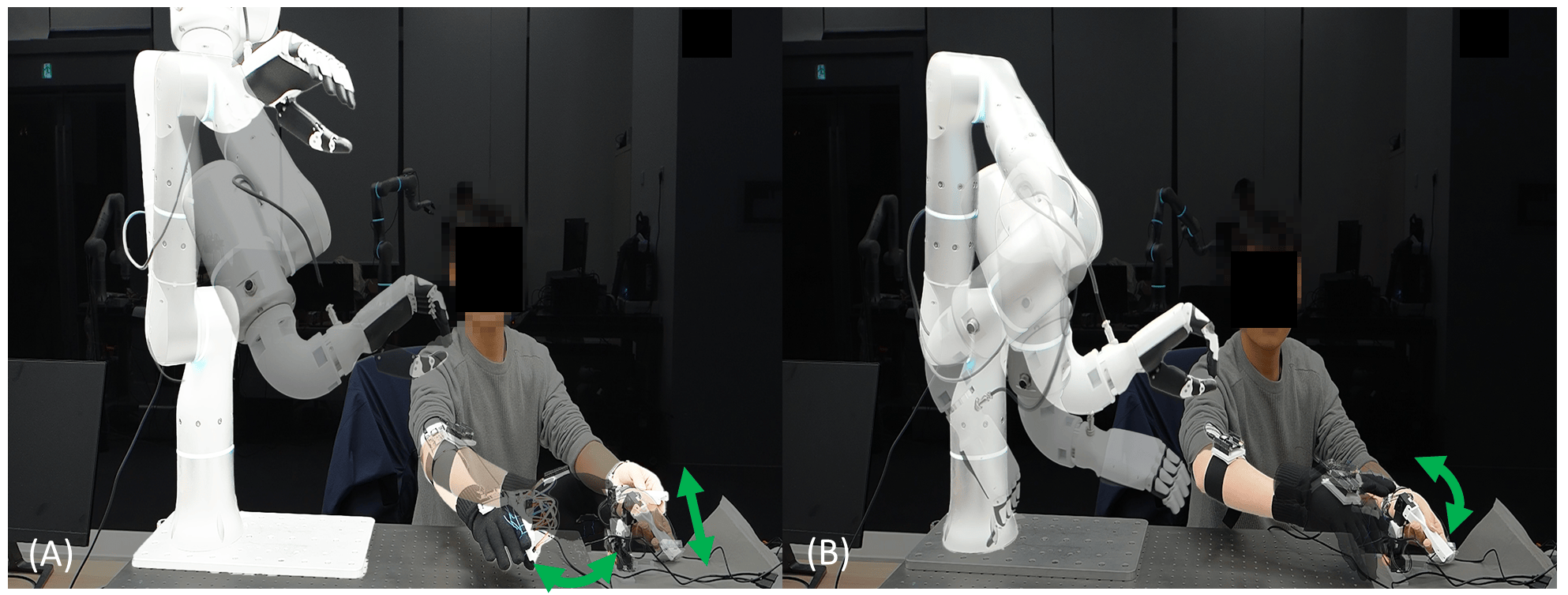}
    \caption{Physical realization of spatial decoupling. (A) the operator moves the global and local components simultaneously. (B) The operator moves the local component only to manipulate the last three joints of the Flexiv robot arm.}
    \label{fig:implement-spatial}
\end{figure}

\section{Implementation Details}
\label{sec:setup}

\begin{figure}[!t]
    \centering
    \includegraphics[width=\linewidth]{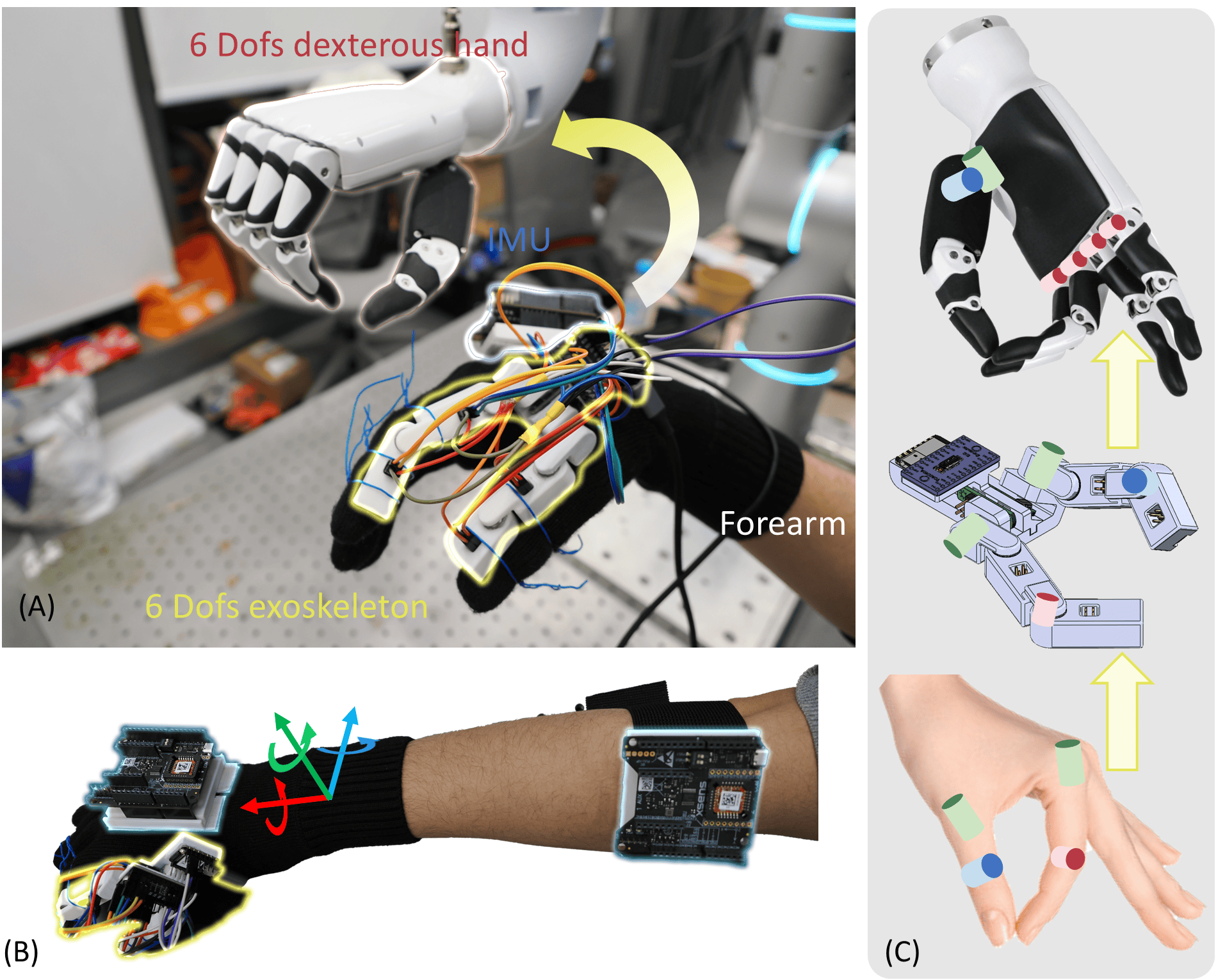}
    \caption{The two-finger exoskeleton design in the spatial decomposition framework. The thumb of the Inspire hand follows the human operator's thumb, and the index, middle, ring, and pinky fingers follow the human operator's index. (B) shows how the two MTI-3 sensors are mounted on the operator's arm for wrist rotation measurement.}
    \label{fig:exoskeleton}
\end{figure}

Figure \ref{fig:implement-temporal} and \ref{fig:implement-spatial} depicts the two teleoperation systems developed to implement the proposed temporal and spatial decoupling framework. The dual-arm teleoperation system, designed according to temporal decoupling, utilizes two Piper Agile X Limit robotic arms as the slave system. For gross manipulation, the global component employs Dynamixel servomotors, while the local component integrates two Touch X haptic devices for enhanced fine control. Each Piper arm, a 6-DOF lightweight manipulator, is equipped with a two-fingered gripper for position-controlled grasping. The master system's global component comprises six Dynamixel servomotors: two XL430 motors for the base joints and four XL330-M288 motors for the remaining degrees of freedom. The Touch X devices, serving as the local component, are 6-DOF haptic interfaces that track the human operator's finger tip position and orientation precisely. Similar to the single-arm system, all devices are managed and coordinated through a multi-process LabVIEW program. 

\begin{figure*}[!t] 
    \centering
    \includegraphics[width=\linewidth]{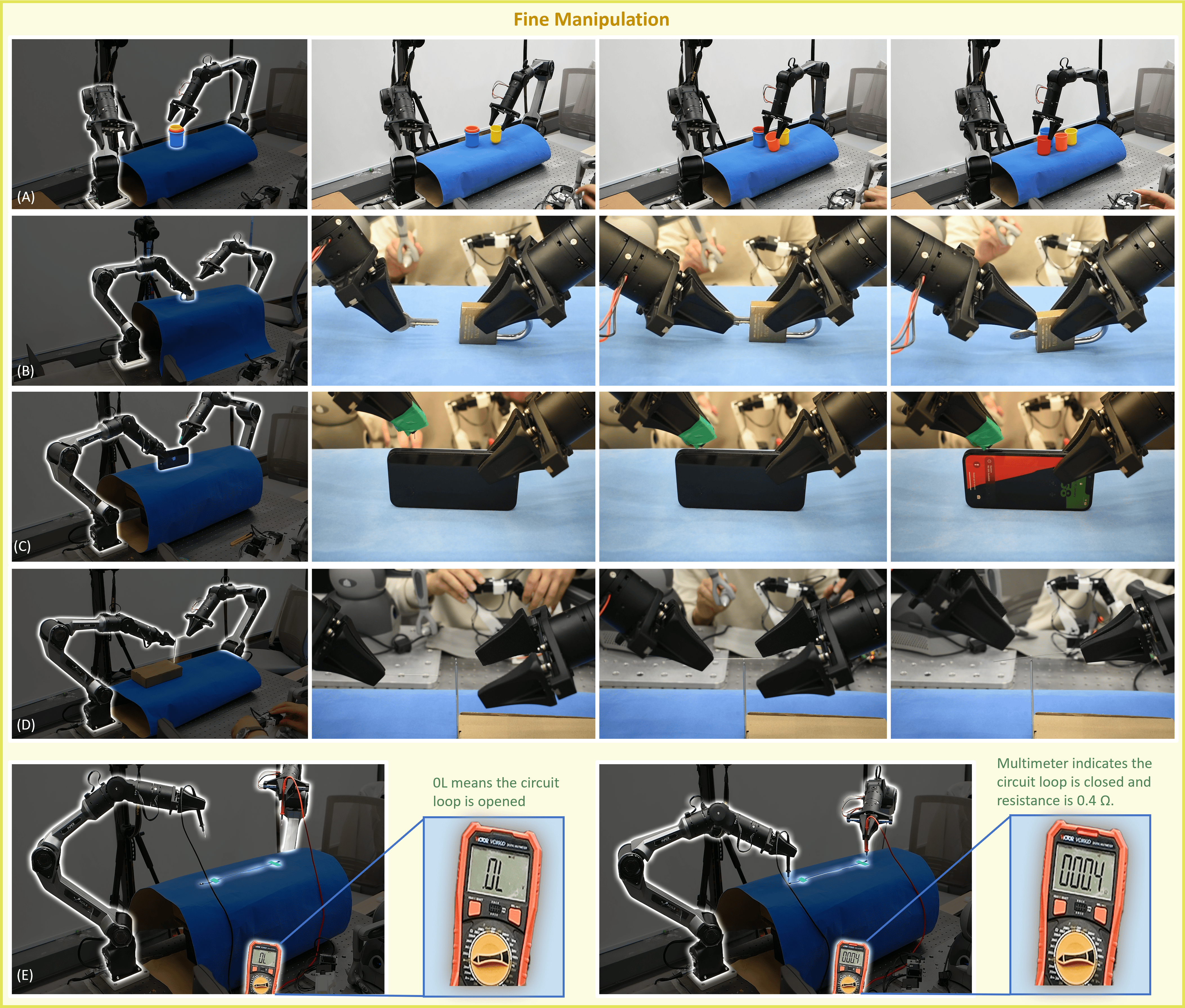}
    \caption{Snapshots of tasks requiring precise manipulation. (A) Taking out Nested Buckets. (B) Key Insertion. (C) SIM Card Needle Insertion. (D) Needle Threading. (E) Wire Testing with Multimeter. The supplementary video of this paper shows the whole process of these demonstrations. Examples (B), (C), (D), and (E) feature bimanual fine manipulation, where both slave arms need to maintain precise position or orientation for completion.}
    \label{fig:precise-demo}
\end{figure*}

The single-arm teleoperation system was developed based on spatial decoupling. Its slave system comprises a Flexiv Rizon robotic arm equipped with an Inspire dexterous hand. The global component of the master device utilizes a Dynamixel servomotor structure to control the robot arm's primary joints, while the local component employs two MTI-3 inertial measurement units (IMUs) and a dexterous exoskeleton for fine manipulation. The Inspire hand, a 6-DOF, five-fingered anthropomorphic hand, is mounted to the Flexiv arm's end-effector with a 90-degree offset to mitigate kinematic singularities. The master system's global component consists of four serially-linked Dynamixel XL330-M288 servomotors, kinematically replicating the Flexiv arm's proximal four joints. For the local component, two MTI-3 IMUs are attached to the operator's forearm and palm, respectively, and a dexterous exoskeleton, integrated with an operational glove, captures index and thumb motions. All devices are interfaced and coordinated through a multi-process LabVIEW program. Upon first launch, the starting orientation of the two IMUs will be respectively recorded as the home orientation, denoted as $R^i_1$ and $R^i_2$. When the local component is activated and the two IMUs report current orientations $R_1$ and $R_2$, we take their rotational displacements ${R^i_1}^{-1}R_1$ and ${R^i_2}^{-1}R_2$. The difference between the forearm and hand IMU, which is $R_s=R_1^{-1}R^i_1{R^i_2}^{-1}R_2$, represents the combined rotation of the last three joints in the Flexiv arm. The last three joint positions are computed by finding the instrinsic X-Y-Z Euler angles that lead to $R_s$.

An exoskeleton is designed for the single-arm teleoperation system to control the Inspire Hand. It is equipped with six AS5600 magnetic encoders and can be attached to the operator's hand to record the thumb and index flexion of the human operator, as shown in Fig. \ref{fig:exoskeleton}. The thumb of Inspire Hand will copy the motion of the operator's thumb, whereas the index, middle, ring, and pinky fingers will imitate the motion of the operator's index following the virtual finger concept \cite{zhou2024dexco, bicchi2011synergies}. The schematics and mapping methods of the exoskeleton will be detailed in the supplementary materials.

\section{Experimental Validations}
\label{sec:experiments}

\begin{figure*}[!t] 
    \centering
    \includegraphics[width=\linewidth]{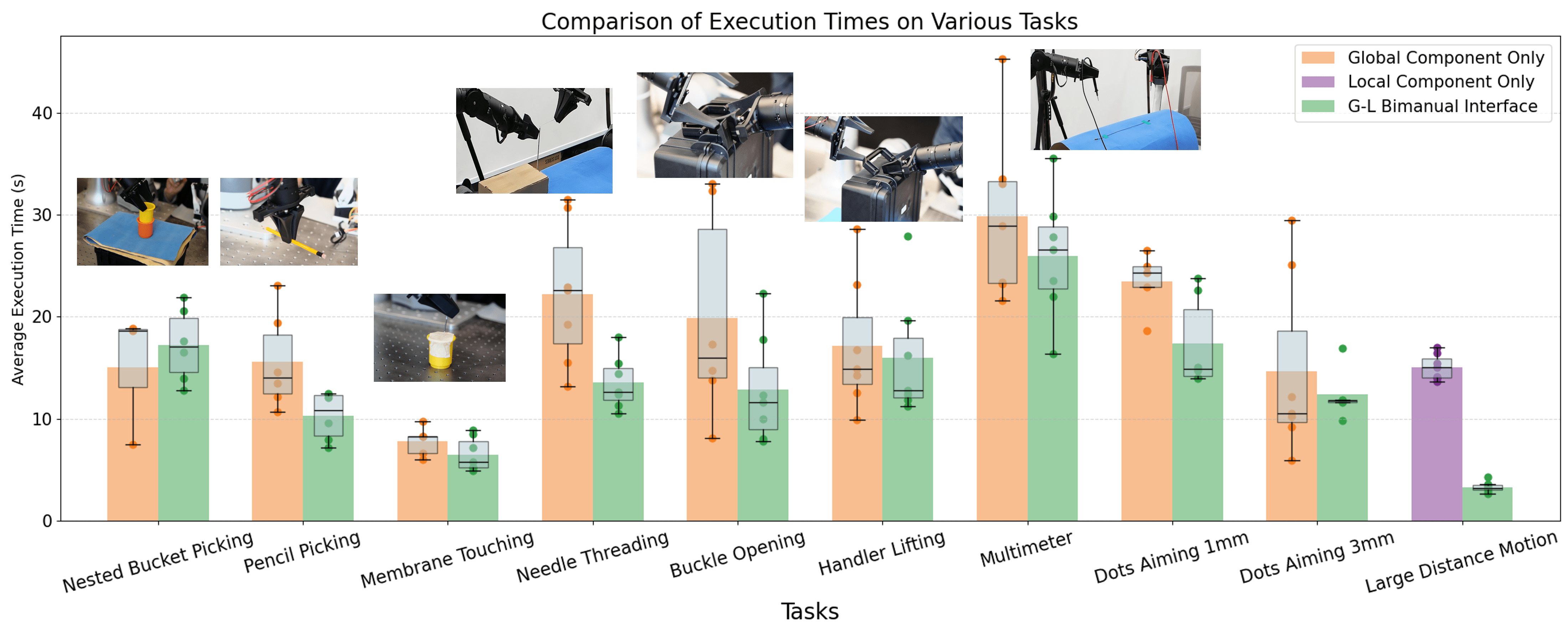}
    \caption{Comparison of completion time between global component only, local component only, and our proposed G-L interface. The G-L interface outperforms only global component in most tasks, except in the Nested Bucket Picking. As shown in Table \ref{tab:exp-succ}, despite taking less time to complete successful trials, only global component has a significantly lower success rate in Nested Bucket Picking. In large distance motion, G-L interface significantly outperforms local component only.}
    \label{fig:exp-time-compara}
\end{figure*}

The performance of the implemented temporal decoupling and spatial decoupling teleoperation systems, mentioned in section \ref{sec:method}, are evaluated through three distinct task categories. The first category comprises tasks with stringent accuracy requirements designed to assess the master system's capacity for precise telemanipulation. The second category encompasses tasks necessitating extensive movement relative to the slave system's workspace, demonstrating the system's ability to effectively utilize the robot's full range of motion. Finally, the third category involves tasks demanding the control of specialized dexterous end-effectors, specifically a five-fingered robotic hand in this study.

\subsection{Precise Manipulation}
\label{subsec:exp-precise}

\begin{figure*}[!t] 
    \centering
    \includegraphics[width=\linewidth]{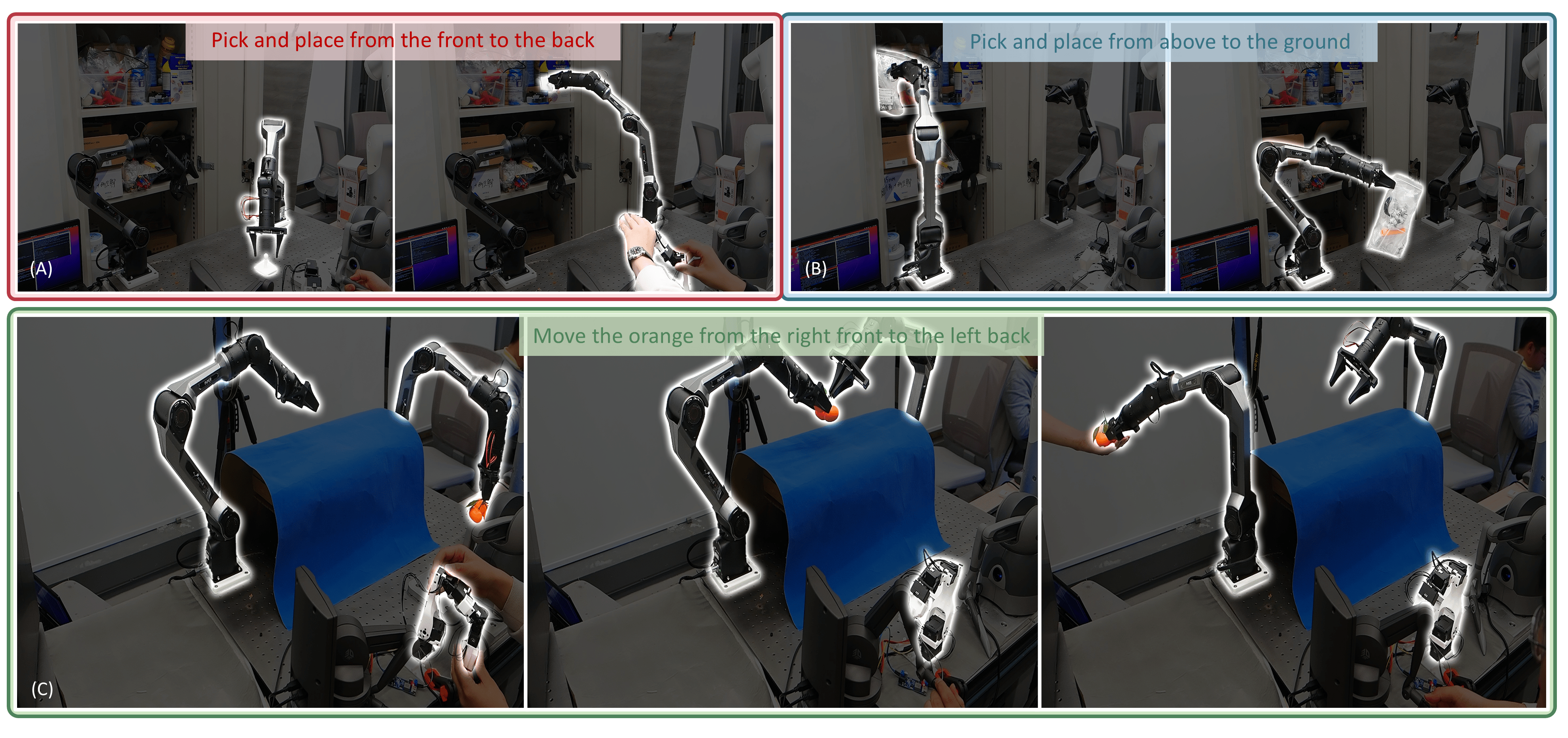}
    \caption{Snapshots of tasks requiring large workspace. (A) Move an object from the desk to the cabinet. (B) Take a bag from the cabinet and place it on the desk. (C) Bimanual handling of oranges from the front-left of the workspace to the back-right. The supplementary video of this paper shows the whole process of these demonstrations.}
    \label{fig:range-demo}
\end{figure*}

To assess the precision and dexterity afforded by the temporal decoupling teleoperation system (dual-arm setup), we conducted a series of experiments involving intricate manipulation tasks. These tasks, detailed below, were selected to represent challenges commonly encountered in real-world scenarios and to highlight the system's capabilities in addressing them. Snapshots of these tasks are shown in Fig. \ref{fig:precise-demo}.

\begin{itemize}
    \item \textbf{Key Insertion}: This task, known for its demanding precision requirements, involves inserting a key into a standard bronze lock. The small tolerances between the key and lock necessitate precise control of both linear and rotational motion, as even minor deviations can result in failure or damage to the lock.
    \item \textbf{Needle Threading}: This bimanual task requires one slave arm to manipulate a fishing line through a needle eye with a radius of approximately 1.5 mm. The second slave arm then grasps the line and pulls it through, demonstrating coordinated manipulation and fine teleoperation.
    \item \textbf{Taking out Nested Buckets}: This task evaluates the system's ability to manipulate objects with complex geometries and contact dynamics. Using a set of four nested buckets from the Yale-CMU-Berkeley Object and Model Set \cite{calli2017yale}, the slave arms must sequentially extract each bucket and place it stably on a surface, avoiding collisions or accidental drops.
    \item \textbf{SIM Card Needle Insertion}: This task replicates a common yet delicate manipulation requiring precise alignment. The slave robot must accurately insert a SIM card needle into the small ejector hole of a mobile phone, then apply the necessary force to release the SIM card tray. This highlights the system's capacity for fine control in constrained environments.
    \item \textbf{Wire Testing with Multimeter}: This task assesses the system's ability to perform precise movements in a functional context. The slave robot uses a multimeter to test the conductivity of a Dupont wire by contacting the probes to the small metallic openings on the wire's connectors. This task underscores the system's suitability for tasks requiring both precision and tool manipulation.
\end{itemize}

To quantitatively evaluate the performance improvements offered by the temporal decoupling teleoperation system, we conducted a comparative study with isolated global and local component. Seven participants were recruited to complete a series of tasks under two conditions: (1) using the global component (equivalent to the GELLO system) or local component only, and (2) using both the global and local components. Each participant performed each task twice, once under each condition. Regrasping and retrying were permitted unless the task was deemed failed according to the predefined criteria detailed below. In the event of failure, the respective trial was disregarded when computing the average completion time. Prior to the experiment, participants received a brief tutorial on operating the teleoperation system and were given one hour of practice time to familiarize themselves with the system and task procedures. For each trial, the master and slave systems were initialized to identical starting configurations, and the completion time was recorded.

Snapshots of the following tasks designed for assessment can be found in the supplementary materials. In addition, the \textbf{Needle Threading} and \textbf{Wire Testing with Multimeter} are also included. As both tasks have no failure criteria, participants made continuous attempts until successful.

\begin{itemize}
    \item \textbf{Nested Bucket Picking}: Two nested bins were placed on a desk. Participants were tasked with lifting the inner bin without disturbing the outer bin. If either bin fell off the desk during the manipulation, the task was considered to be failed.
    \item \textbf{Pencil Picking}: Participants were tasked with picking up a pencil from a random location on a desk using the slave robot. Contacting the desk with excessive force using the robot gripper resulted in task failure. The pencil's position was randomized for each trial, but participants were allowed to practice with a pencil beforehand.
    \item \textbf{Membrane Touching}: A thin piece of scotch tape, simulating a delicate membrane, was placed over the opening of a small cup. Participants were instructed to bring the needle held by the slave robot into contact with the membrane without puncturing it. Penetration constituted task failure.
    \item \textbf{Buckle Opening}: Participants were asked to open a buckle on a box to open it using the slave robot. Since this task involves close and tight contact with an object, emergency stop of the Piper robot can be triggered if the joint torque exceeds its safety limit. The task was considered failed if the emergency stop is triggered.
    \item \textbf{Handle Lifting}: Given a box whose handle is initially laid down, participants are required to rotate the handle to an upright pose and lift the box using the handle. The failure condition is the same as the \textbf{Buckle Opening} task.
    \item \textbf{Dots Aiming}: This task is an abstraction of many accurate insertion tasks like SIM card needles, shafts, and TRS connectors. Three dots with radius 1 mm, 2 mm, and 3 mm are printed on a target sheet. The slave system pinches a needle with its gripper. Volunteers are asked to prick each dot with this needle using the master system. Participants cannot access the test time target sheet position, but they are allowed to practice with the same target sheet before the test starts. This task is not given a failure criteria so participants can make continuous attempts until they are successful.
\end{itemize}

\begin{figure*}[!t] 
    \centering
    \includegraphics[width=\linewidth]{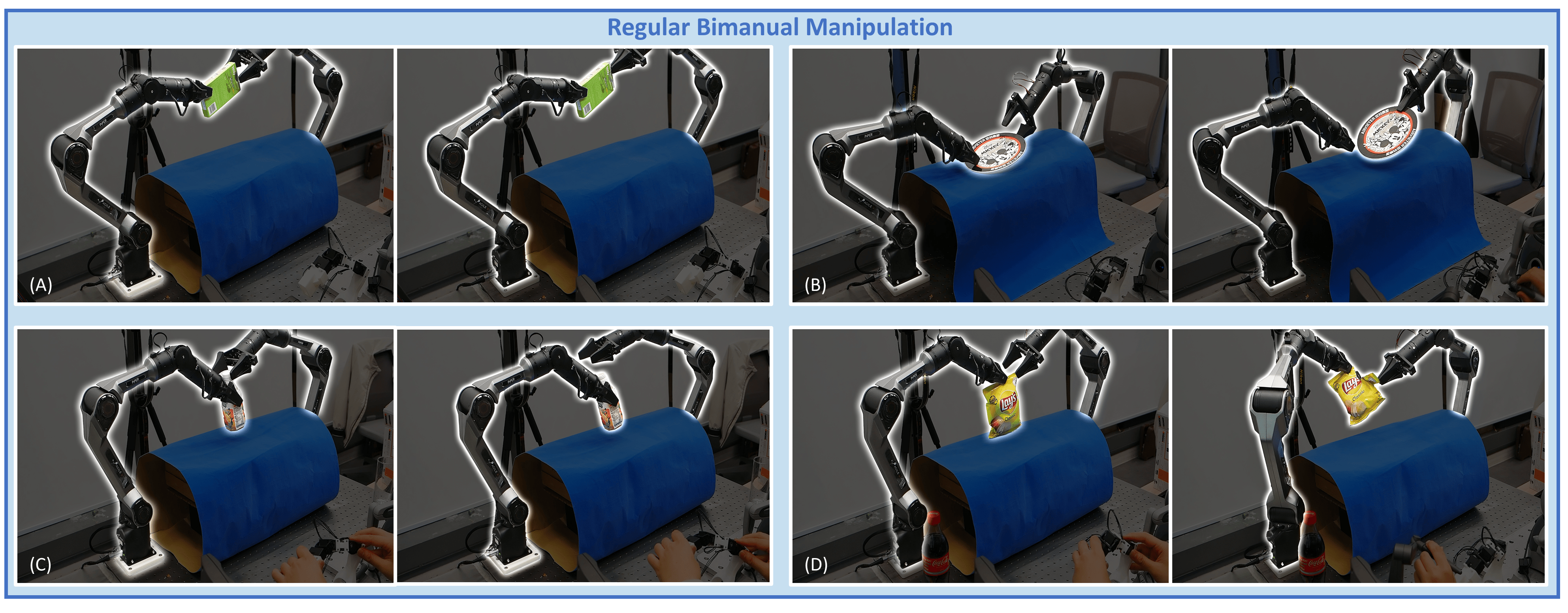}
    \caption{Snapshots of daily object manipulation. (A) Opening a snack box. (B) Dual arm plate grasping. (C) Opening bottle cap. (D) Opening chip bag.}
    \label{fig:bimanual-demo}
\end{figure*}

Fig. \ref{fig:exp-time-compara} presents the results of the quantitative study, demonstrating that the G–L interface consistently outperforms the isolated global or local components across most tasks. While participants using all configurations were able to quickly move the slave robot’s end-effector near the target area, those operating with the global-only setup frequently struggled to perform the final fine motions required for task completion. In contrast, the local component within the G–L interface enabled smoother and more efficient fine-tuning, resulting in significantly shorter overall completion times. The isolated global component system demonstrated marginally faster completion in nested bucket picking. However, it has a considerably lower success rate comparing to the G-L interface, as elaborated in Table \ref{tab:exp-succ}.

\begin{table}[!t]
    \centering
    \begin{tabular}{||c | c | c||}
       \hline
                       & Global Component & G-L Interface \\ [0.5ex] 
         \hline\hline
         Nested Bucket Picking & $3/7$ & $6/7$ \\ 
         \hline
         Pencil Picking     & $7/7$ & $7/7$ \\
         \hline
         Membrane Touching  & $5/7$ & $7/7$ \\
         \hline
         Buckle Opening  & $6/7$ & $7/7$ \\
         \hline
         Handle Lifting  & $7/7$ & $7/7$ \\
       \hline
    \end{tabular}
    \caption{Success rate of isolated global component and G-L interface on each task. Only tasks with pre-defined failure conditions in section \ref{subsec:exp-precise} are included. The G-L interface shows a higher success rate }
    \label{tab:exp-succ}
\end{table}

\subsection{Large Range Manipulation}
\label{subsec:exp-large}

\begin{figure*}[!t] 
    \centering
    \includegraphics[width=\linewidth]{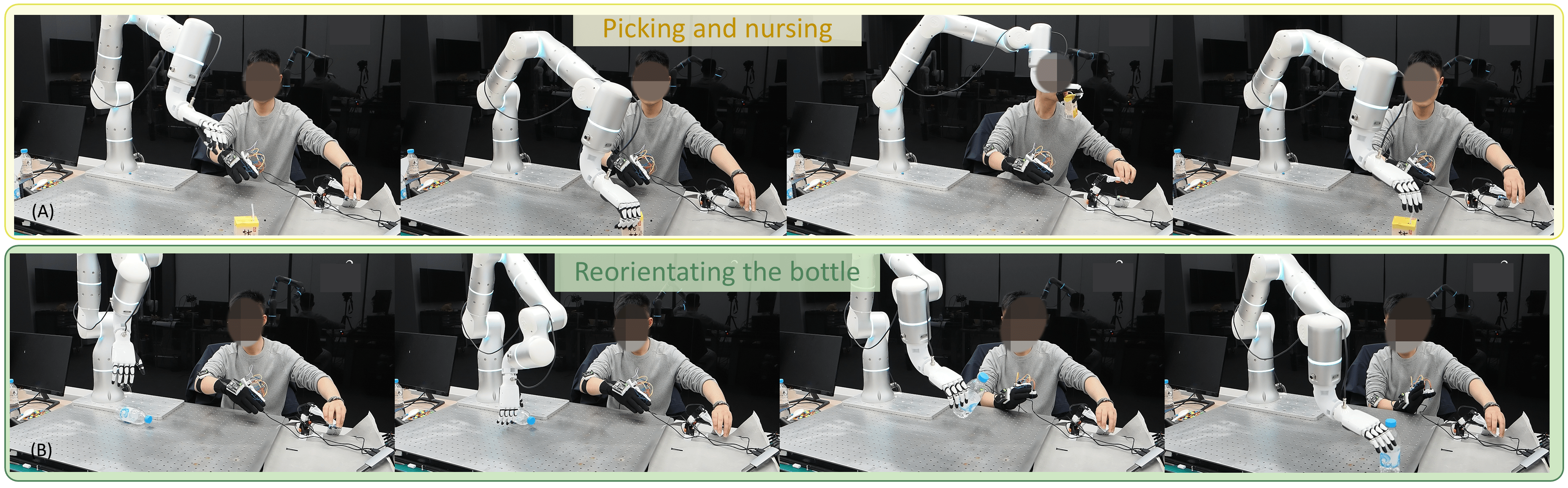}
    \caption{Snapshots of tasks with dexterous manipulation. (A) The operator needs to fetch a box of drink from a distance away, take a sip and then put the drink box back. (B) Reorientating a water bottle from the desk.}
    \label{fig:dexterous-demo}
\end{figure*}

To assess the ability of the G-L interface to effectively utilize the entire workspace of the slave robot, we conducted two demonstrations involving large-scale movements spanning the workspace's full range along the left-right, front-back, and vertical axes. These demonstrations aimed to confirm that the integration of precise manipulation capabilities within the G-L interface did not compromise its capacity for executing large-range manipulation tasks.

\begin{itemize}
    \item \textbf{Object Handling}: This demonstration involved a coordinated bimanual task in which one slave robot retrieved an orange from the far left-front corner of its workspace and transferred it to the second slave robot. The second robot then handed off the orange to a human operator located at the far right-back corner of the workspace. This task exemplifies the system's ability to facilitate long-range object transfer and cooperative manipulation, similar to demonstrations in prior teleoperation systems \cite{wu2023gello}.
    \item \textbf{Pick and Place from a Cabinet}: This demonstration involved interacting with a tall cabinet positioned behind the slave robot. The task required retrieving specific objects from a cluttered arrangement within the cabinet and placing them in front of the robot, or conversely, placing objects from the workspace into the cabinet. This task highlights the system's capacity for navigating and manipulating objects within complex, spatially extended environments.
\end{itemize}

Both tasks are completed successfully and their snapshots are shown in Fig. \ref{fig:range-demo}. These tasks are good examples of how global and local components can cooperate to improve user experiences. The global component ensures a large range of motion while the local component enables the operator to make fine adjustments in clustered environments.

\subsection{Daily Object Manipulation}
\label{subsec:exp-daily}

Tasks demonstrated in sections \ref{subsec:exp-precise} and \ref{subsec:exp-large} are mainly achieved via single-arm manipulation. In this section, we show that the temporal decoupling teleoperation system can also complement bimanual teleoperation tasks involving daily objects. These tasks include picking up a plate, where one slave robot must push down one edge to lift up the other, thereby allow another slave robot to pick up the entire plate. Tasks like opening a chip bag, snack box, and water bottle are also included.

\subsection{Dexterous Manipulation}
\label{subsec:exp-dex}

This section details the master system built from spatial decoupling that is capable of controlling a complex end-effector. We use two tasks to demonstrate its dexterity: Feeding the human operator a drink and reorientating a water bottle. Snapshots of these tasks are shown in Fig. \ref{fig:dexterous-demo}.

\section{Discussions and Limitations}
\label{sec:discuss}

Despite its advantages, the proposed G-L interface presents certain limitations, primarily stemming from its design and hardware implementation. These limitations, outlined below, warrant consideration for future development and refinement.

\begin{itemize}
    \item \textbf{Additional Hardware Costs}: Unlike conventional teleoperation systems that typically employ a single master device per slave robot, the G-L interface often requires two sets of master devices. While this configuration contributes to enhanced performance, it also leads to increased hardware costs.
    \item \textbf{Bimanual Control for Unimanual Tasks}: The temporal decoupling method inherent to the G-L interface allows for straightforward bimanual teleoperation by preventing simultaneous activation of the global and local components. However, in the spatial decoupling method, where both components are engaged concurrently, the operator must utilize both hands to control a single slave robot, potentially increasing operational complexity.
    \item \textbf{Challenges in Orientation Control}: Empirical observations revealed that fine-tuning the orientation of the slave robot's end-effector proved more challenging than adjusting its position. Within the temporal decoupling framework, the local component's reliance on scaled motion tracking from the touch-X device limits its efficacy to small orientation adjustments. Larger adjustments can become unintuitive due to the scaled mapping between master and slave rotations. Conversely, while the dual-IMU design in the spatial decoupling framework permits larger orientation adjustments, it introduces sensitivity to unintended skin movement as the operator twists their arm. These movements can increase discrepancies between the measured wrist rotation and its ground truth, leading to counterintuitive behavior in the slave end-effector's orientation. This issue could potentially be mitigated by incorporating a more robust and precise wrist rotation sensing mechanism.
\end{itemize}

Interestingly, a notable learning curve was observed among participants. Initial trials were often characterized by cautious and slow movements, with proficiency and speed increasing with practice. To account for this learning effect and ensure a fair comparison, each participant was provided with one hour of dedicated practice time prior to data collection.

\section{Conclusion}
\label{sec:conclusion}

This paper introduced the G-L interface for on-demand teleoperation, a novel approach that decouples the master system into global and local components responsible for large-range movements and fine manipulations, respectively. This design enables the master system to effectively make precise adjustments while retaining the capacity for utilizing the entire workspace of the slave robot. The modular nature of the G-L interface allows for flexible implementation tailored to specific task requirements.

This work presented two realizations of the G-L interface: temporal decoupling and spatial decoupling. Through demonstrations and experimental evaluation, we have shown that temporal decoupling facilitates tasks demanding high accuracy while maintaining full workspace utilization. Conversely, spatial decoupling enables the teleoperation of complex end-effectors with enhanced dexterity. These findings highlight the versatility and potential of the G-L interface to advance the capabilities of teleoperation systems across a wide range of applications.

\bibliographystyle{./bib/IEEEtran}
\bibliography{./bib/main}

\newpage

\appendix

The following information will be included in this section.

\begin{itemize}
    \item Design details of the exoskeleton in section \ref{subsec:spatial-decoup}, especially how human operator's hand motion is mapped to the dexterous hand's motion.
    \item Snapshots of Tasks used for quantitative studies, which data in Fig. \ref{fig:exp-time-compara} is collected on.
\end{itemize}

We have also attached supplementary videos showcasing the mechanisms of the temporal and spatial decoupling, and the tasks in Fig. \ref{fig:precise-demo}, \ref{fig:range-demo}, \ref{fig:bimanual-demo}, \ref{fig:dexterous-demo} of the paper.

\textbf{Appendix A: Exoskeleton Design}

\begin{figure}
    \centering
    \includegraphics[width=0.8\linewidth]{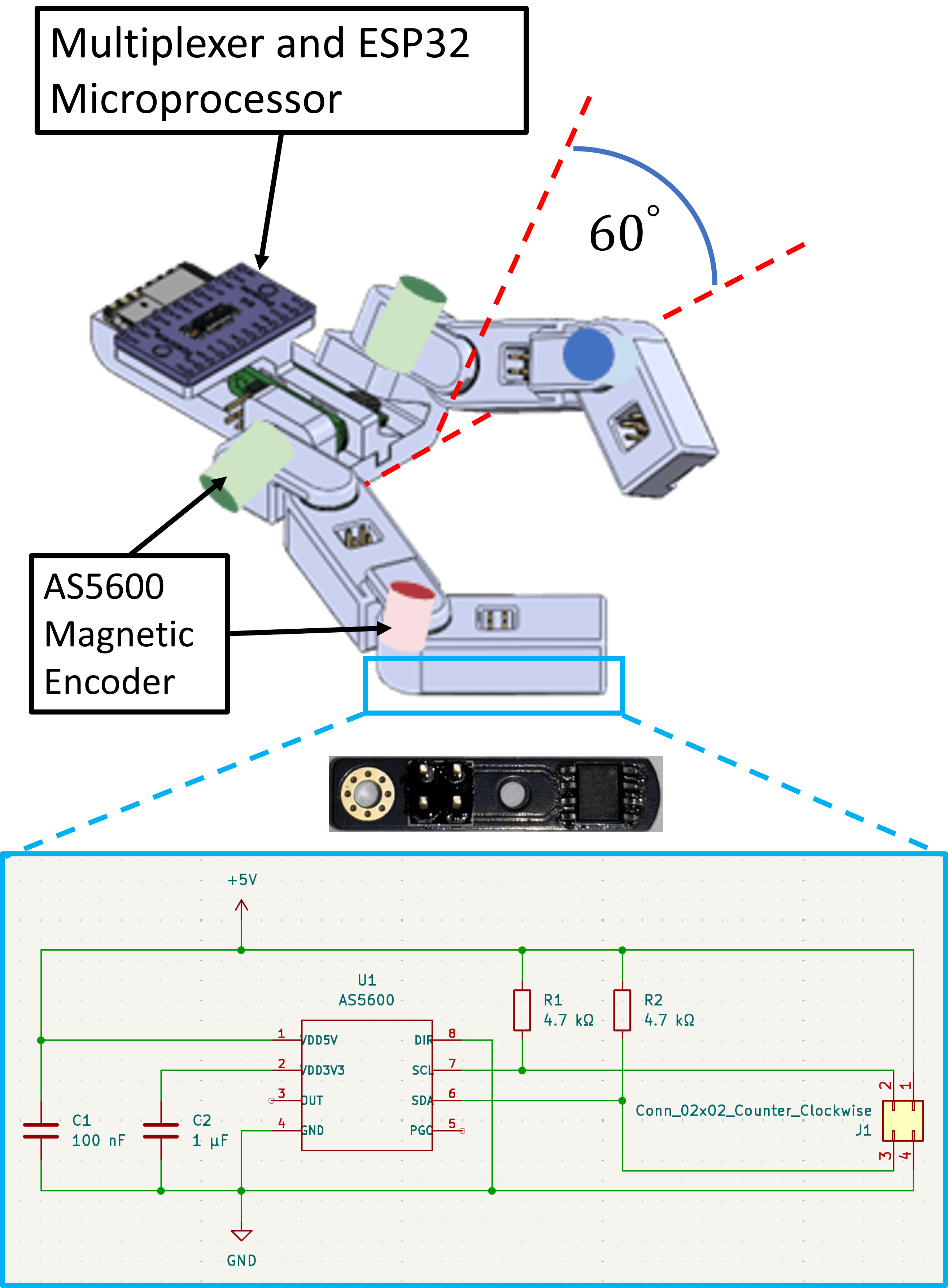}
    \caption{Design details of the exoskeleton master that controls the Inspire dexterous hand.}
    \label{fig:exo-design}
\end{figure}

For better fit with the human operator's hand, we added a $60^\circ$ tilting angle to the thumb of the exoskeleton. Fingers on the exoskeleton have two joints, each equipped with an AS5600 magnetic encoder. We designed a circuit to send encoder readings to a multiplexer via I2C protocol. The multiplexer is then connected to an ESP32 microprocessor, which allows communication with a computer. Fig. \ref{fig:exo-design} further elaborates the exoskeleton's structural design and the communication circuit.

\begin{figure}
    \centering
    \includegraphics[width=\linewidth]{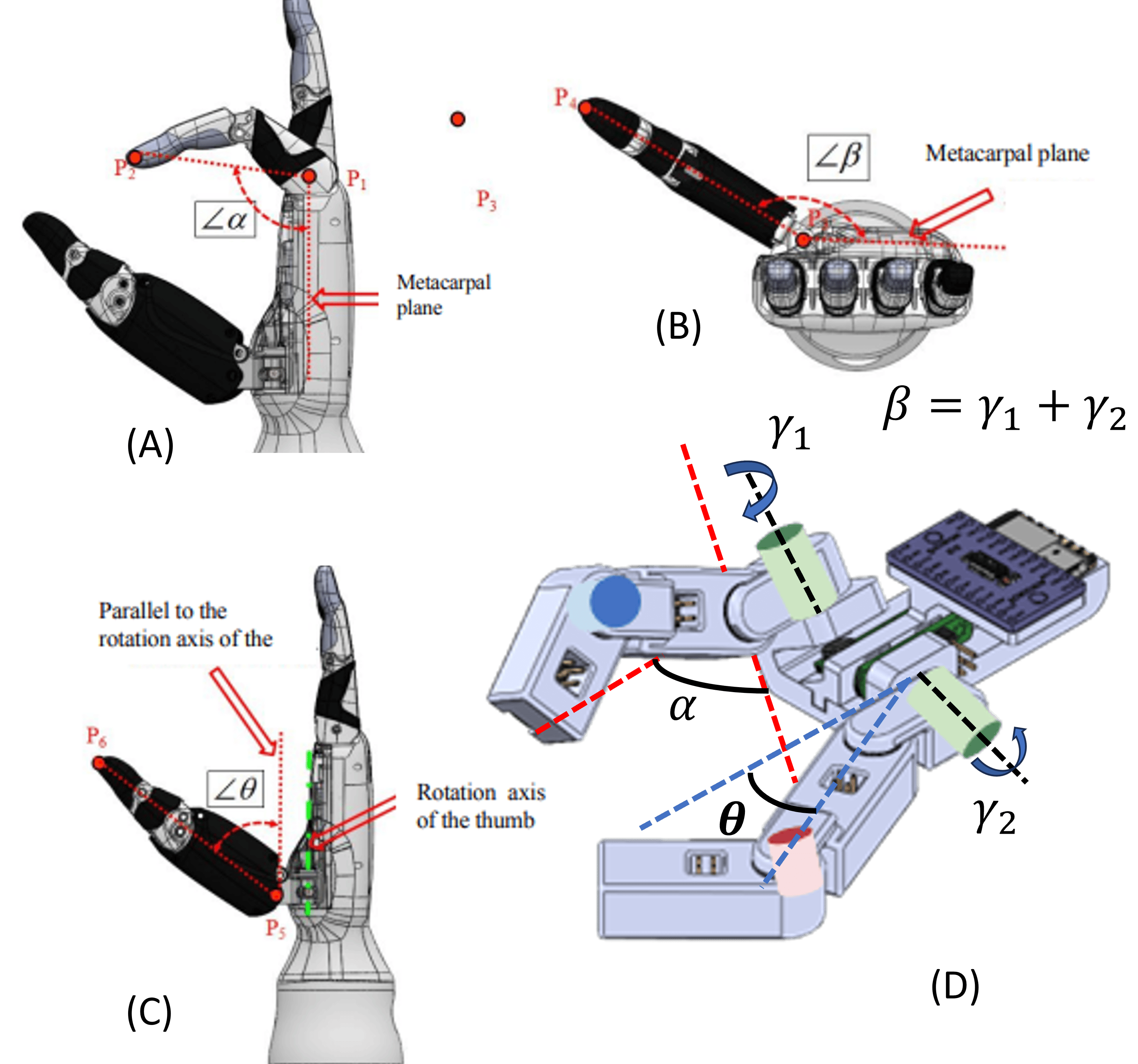}
    \caption{Mapping between the exoskeleton and Inspire dexterous hand's state. Subfigures (A), (B) and (C) are duplicated from the manual of the Inspire hand.}
    \label{fig:joint-mapping}
\end{figure}

The Inspire dexterous hand is cable-driven with 6 DoFs in total, two on the thumb and one for each of the remaining fingers. Its position state is represented by the bending and rotational angles between the fingertip and the metacarpal plane. Therefore, while mapping the exoskeleton's position state to the Inspire hand's position state, we also compute the angle between the fingertip and metacarpal plane of the exoskeleton, as shown in Fig. \ref{fig:joint-mapping}.

\textbf{Appendix B: Further Details of Tasks in Fig. \ref{fig:exp-time-compara}}

This part includes snapshots of tasks in Fig. \ref{fig:exp-time-compara} of the paper for better understanding of the testing criteria. Tasks that have already appeared in Fig. \ref{fig:precise-demo} and the dots aiming tasks are not included here.


\begin{figure*}
\centering
    \begin{subfigure}{0.9\textwidth}
        \includegraphics[width=\linewidth]{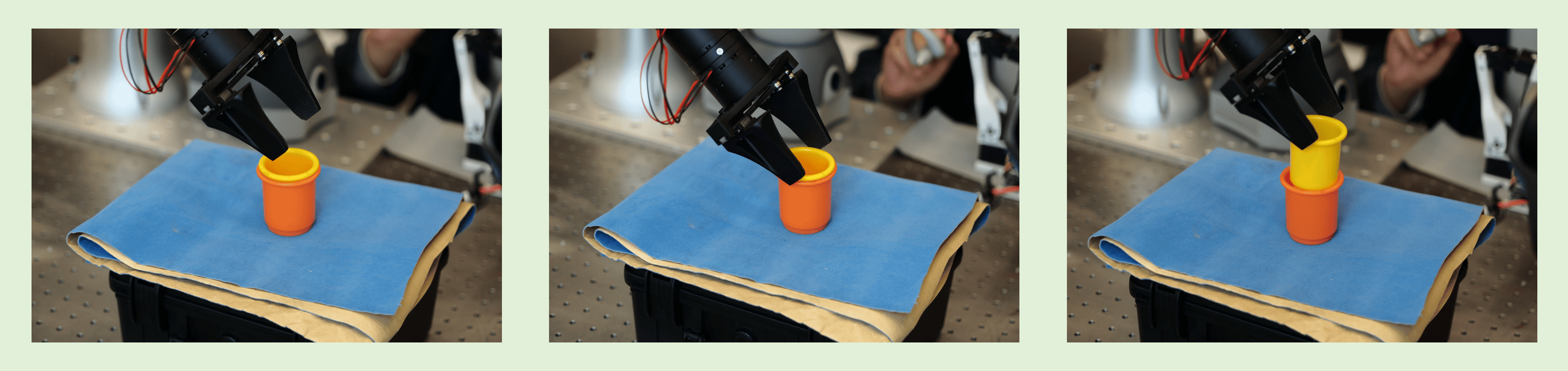}
        \caption{\textbf{Nested Bucket Picking}: Taking out the smaller bucket from the larger bucket. Reattempts commonly happens when both buckets are taken up, or the larger bucket is tipped over when lifting up the smaller bucket. Reattempts usually cause an uncontrollable motion of the larger bucket, and the task is considered failed if the large bucket fall off the platform.}
        \label{fig:nested-bucket-picking}
    \end{subfigure}

    \begin{subfigure}{0.9\textwidth}
        \includegraphics[width=\linewidth]{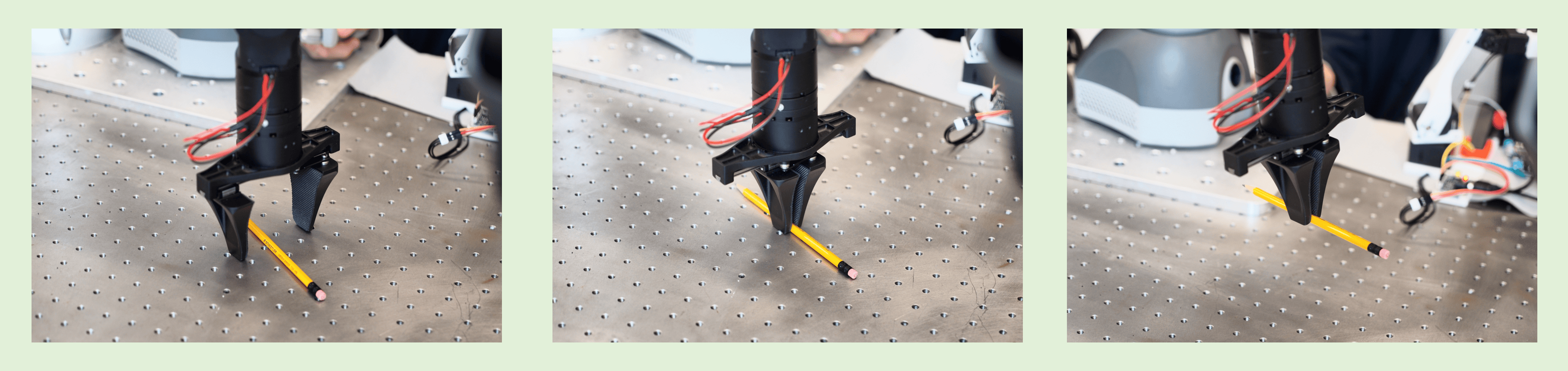}
        \caption{\textbf{Pencil Picking}: If the gripper hits the desk too hard, emergency stop will be triggered and the task is considered failed.}
        \label{fig:pencil-picking}
    \end{subfigure}

    \begin{subfigure}{0.9\textwidth}
        \includegraphics[width=\linewidth]{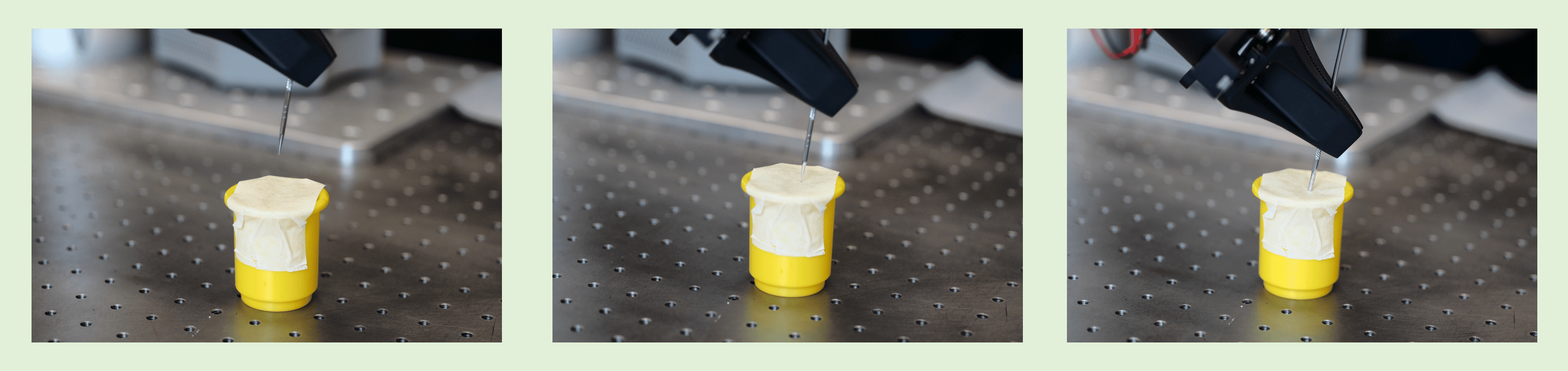}
        \caption{\textbf{Membrane Touching}: The middle subfigure is a succeed case where the needle is touching the scotch tape. The right subfigure is a failure case where the needle penetrates through the scotch tape.}
        \label{fig:membrane-touching}
    \end{subfigure}

    \begin{subfigure}{0.9\textwidth}
        \includegraphics[width=\linewidth]{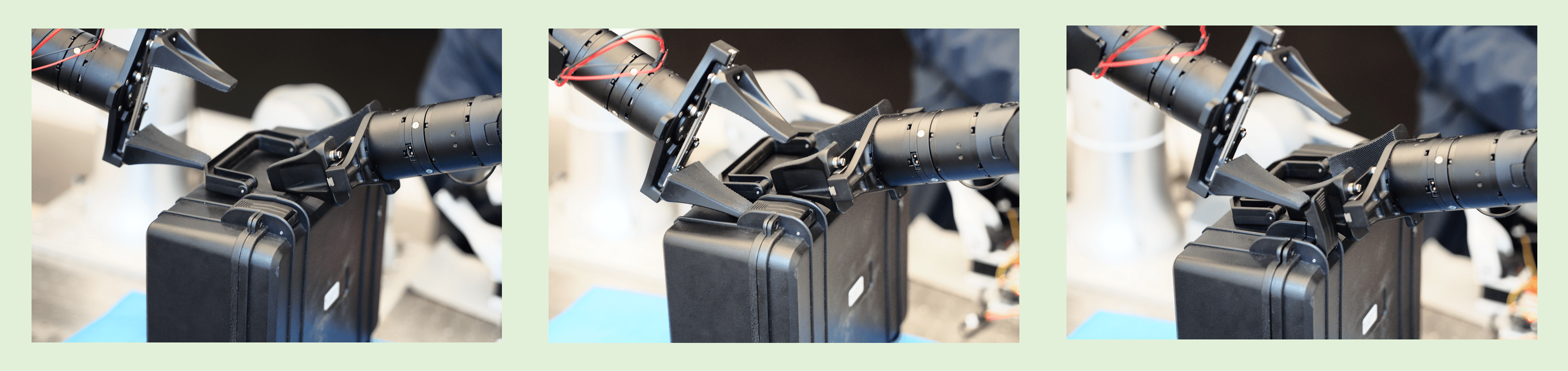}
        \caption{\textbf{Buckle Opening}: If the gripper pushes the box too hard in a wrong direction, emergency stop will be triggered and the task is considered failed.}
        \label{fig:buckle-opening}
    \end{subfigure}

    \begin{subfigure}{0.9\textwidth}
        \centering
        \includegraphics[width=\linewidth]{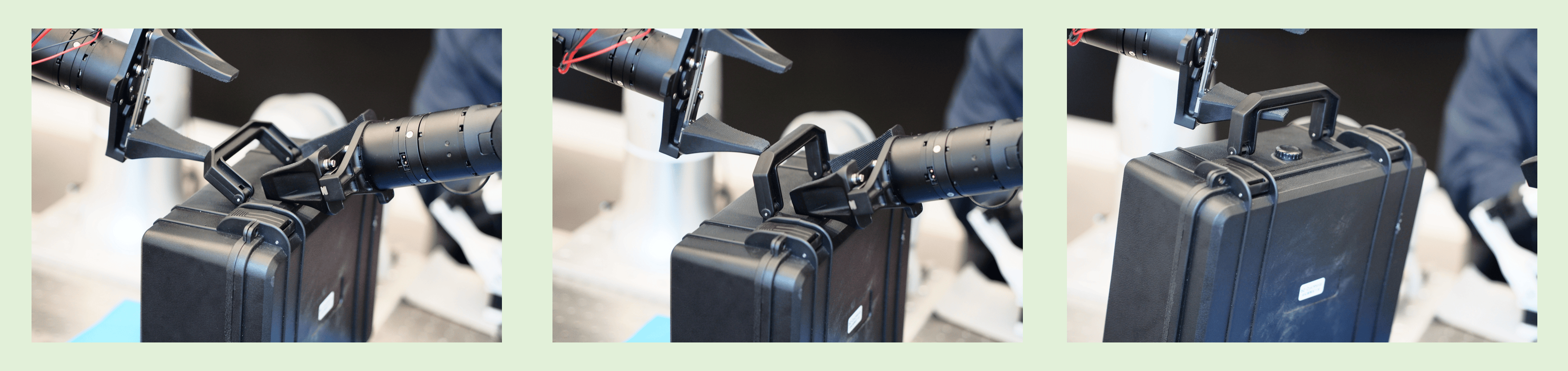}
        \caption{\textbf{Handle Lifting}: the slave robot must rotate the and and lift up the box to be considered success.}
        \label{fig:handle-lifting}
    \end{subfigure}

\caption{Snapshots and further explanations of tasks for quantitative studies in Fig. 8.}
\label{fig:quant_tasks}
\end{figure*}

\end{document}